%% file: main.tex
\documentclass[journal]{IEEEtran}

\usepackage{xcolor,soul,framed} 

\colorlet{shadecolor}{yellow}
\usepackage[pdftex]{graphicx}
\graphicspath{{../pdf/}{../jpeg/}}
\DeclareGraphicsExtensions{.pdf,.jpeg,.png}

\usepackage[cmex10]{amsmath}
\usepackage{bm}

\usepackage{array}
\usepackage{mdwmath}
\usepackage{mdwtab}
\usepackage{eqparbox}
\usepackage{url}
\usepackage{caption}  
\usepackage{pifont}
\usepackage{multirow}
\usepackage{booktabs}
\newcommand{\cmark}{\ding{51}}%
\usepackage{amsmath}
\usepackage{amssymb}

\begin{document}
\bstctlcite{IEEEexample:BSTcontrol}
    \title{CoNav Chair: Development and Evaluation of a Shared Control-based Wheelchair for the Built Environment}
  \author{Yifan Xu,
      Qianwei Wang,
      Jordan Lillie,
      Vineet Kamat,
      Carol Menassa,
      and Clive D'Souza

  \thanks{Yifan Xu, Vineet Kamat and Carol Menassa are with the Department of Civil Engineering, University of Michigan, Ann Arbor, MI, 48105 USA (e-mail: \{yfx, vkamat, menassa\}@umich.edu).}
  \thanks{Qianwei Wang is with College of Literature, Science, and the Arts, University of Michigan, Ann Arbor, MI, USA, (e-mail: qweiw@umich.edu.)}
  \thanks{Jordan Lillie is with Wheelchair Seating Service, University of Michigan Health, Ann Arbor, MI, USA, (e-mail: jlillie@med.umich.edu)}
    \thanks{Clive D'Souza is with Department of Rehabilitation Science and Technology, University of Pittsburgh, Pittsburgh, PA, USA, (e-mail: crd85@pitt.edu)}}


\maketitle

\begin{abstract}
As the global population of people with disabilities (PWD) continues to grow, so will the need for mobility solutions that promote independent living and social integration. Wheelchairs are vital for the mobility of PWD in both indoor and outdoor environments. The current state-of-the-art in powered wheelchairs are based on either manually controlled or fully autonomous modes of operation, offering limited flexibility and often proving difficult to navigate in spatially constrained environments. Moreover, research on robotic wheelchairs has focused predominantly on complete autonomy or improved manual control; approaches that can compromise efficiency and user trust. To overcome these challenges, this paper introduces the CoNav Chair—a smart wheelchair based on the Robot Operating System (ROS) and features shared control navigation and obstacle avoidance capabilities that are intended to enhance navigational efficiency, safety, and ease fo use for the user. The paper outlines the CoNav Chair’s design and presents a preliminary usability evaluation comparing three distinct navigation modes, namely, manual, shared, and fully autonomous, conducted with 21 healthy, unimpaired participants traversing an indoor building environment. Study findings indicated that the shared control navigation framework had significantly fewer collisions and performed comparably if not superior to the autonomous and manual modes on task completion time, trajectory length and smoothness; and was perceived as being safer and more efficient based on user-reported subjective assessments of usability. Overall, the CoNav system demonstrated acceptable safety and performance laying the foundation for subsequent usability testing with end users, namely, PWDs who rely on a powered wheelchair for mobility. 
\end{abstract}


%
\IEEEpeerreviewmaketitle


\input{Introduction}
\input{RelatedWork}

\input{Design}
\input{Usability}
\input{Result}
\input{Conclusion}
\input{data}
\input{Acknowledge}


%





\ifCLASSOPTIONcaptionsoff
  \newpage
\fi





\bibliographystyle{IEEEtran}
\bibliography{IEEEabrv,Bibliography}

\vfill


\end{document}

%% file: Introduction.tex
\section{Introduction}
\label{sec:intro}

With the number of people with disabilities (PWD) increasing annually, an expanding group of individuals will have to rely on wheelchairs to facilitate their mobility in the built environment. Mobility impairments among certain PWD restrict travel and their access to education, healthcare, and employment; intensifying physical and mental stress leading to additional chronic conditions and burden on families, caregivers, healthcare systems, and society~\cite{Chutke2022,Gudlavalleti2018}. Wheelchairs, both manual wheelchairs and increasingly electric powered wheelchairs (PWC) are the most common means of assisting PWD navigate indoor and outdoor environments and support activities of daily living and social participation~\cite{Kairy2014,Cooper1998Wheelchair,Sahoo2023}. Of the nearly 25.5 million people in US experiencing mobility challenges, an estimated 5.5 million community-dwelling adults rely on wheelchairs for daily mobility~\cite{taylor2018disability}. The number of adults using wheelchairs is expected to increase significantly due to improvements in post-trauma survivorship (with some physical impairments) and longevity~\cite{King2013,Nie2024}. 

In recent decades, the introduction of smart wheelchairs has shown significant potential to improve independent mobility and the quality of life for PWD~\cite{Fehr2000Adequacy}. Smart wheelchairs—also referred to as intelligent, autonomous, or semi-autonomous wheelchairs—are mobility devices enhanced with sensors, computation, and control algorithms to automate certain aspects of PWC operation and assist users perform tasks such as navigation and obstacle avoidance more effectively. Current smart wheelchairs face a notable limitation: they often focus exclusively on either full autonomy—where navigation occurs without human involvement—or fully manual control, which requires the user to operate the device using a joystick~\cite{Fehr2000Adequacy,Udupa2021}. This dichotomy in the level of automation can result in inefficiencies during navigation and a lack of trust in the technology—critical factors that influence the successful adoption of smart wheelchairs~\cite{McKenna2023}. For example, fully autonomous systems may make unsafe decisions in dynamic environments, while fully manual control can overwhelm users having limited upper extremity strength or dexterity by requiring constant joystick operation. Consequently, these challenges hinder user acceptance and integration into daily life~\cite{Parikh2007}. Furthermore, the limited field of view of sensors and the suboptimal performance of autonomous navigation algorithms in avoiding obstacles can leave users feeling unsafe and less willing to give up control~\cite{Xu2024Socially}. 

\begin{figure}[t]
    \centering
    \includegraphics[width =\columnwidth]{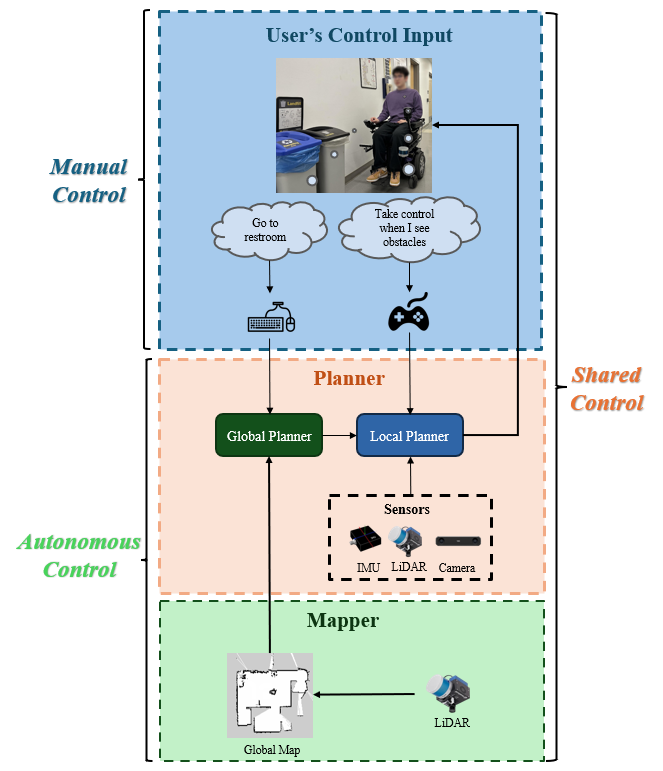}
    \caption{Shared Control Framework}
    \label{fig:General Framework}
\end{figure}

To address these concerns, a shared control-based smart wheelchair offers a potential solution. By combining aspects of both manual control and autonomous control, a shared control device enables users to contribute to decision-making during navigation by partially overriding certain aspects of the autonomous navigation system (Figure~\ref{fig:General Framework}). In the shared control framework, the autonomous system primarily handles path planning and navigation using sensor data (e.g., LiDAR, inertial measurement unit (IMU), camera) to build a map and compute trajectories through the environment. The user’s input serves as corrective feedback, allowing them to intervene when necessary, for example, to adjust direction when encountering obstacles or expressing intent (e.g., “go to the restroom”). This cooperation enables users to retain agency while reducing the cognitive and physical load of continuous manual control. We expect that this approach not only enhances the mobility of PWD but also fosters user trust by offering a balanced partnership between autonomy and manual control.

The objective of this paper is to present the hardware and software aspects for a shared control-based smart wheelchair called the CoNav Chair developed by our team and by leveragin insights from previous smart wheelchair research~\cite{Bourhis1996Mobile,Gribble1998Integrating,Yanco1998Wheelesley,Levine1999NavChair,Park2017DiscreteTime,Kopun2023Toronto}. The proposed system builds on a commercially available PWC platform enhanced with essential sensors to enable intelligent navigation including LiDAR, a video camera, an inertial measurement unit (IMU), and wheel encoders. A closed-loop PID controller~\cite{KiamHeongAng2005} ensures stable motor control. On the software side, a ROS-based shared control framework implementation integrates user joystick input with autonomous planning to support adaptive navigation in dynamic indoor environments. This approach enables responsive obstacle avoidance, social awareness, and user preference alignment~\cite{Xu2024Socially}.

To assess the technical performance of the proposed shared control wheelchair system, we conducted a preliminary usability study in a controlled indoor environment. Twenty-one participants navigated a narrow building corridor using the CoNav Chair in three different modes of operation: shared control, fully autonomous (via Model Predictive Control), and conventional manual joystick control. These modes represent a spectrum of interaction from full user control to full autonomy. Following the approach by~\cite{Kutbi2021}, we evaluated both objective measures (e.g., task completion time, trajectory length) and subjective measures (i.e., self-reported ratings) of safety, efficiency, and ease of use to obtain a comprehensive understanding of how the different control modes influence usability and system performance.

\subsection{Statement of Contributions}

This paper makes three important contributions.
\begin{itemize}
    \item We present a novel shared control-based smart wheelchair platform called the CoNav Chair, which using a shared control algorithm bridges the gap between full autonomy and manual control, and thereby advances assistive mobility solutions.
    \item We design and conduct a controlled study with 21 healthy participants without impairment to evaluate and compare the usability (efficiency, safety and ease of use) of three distinct modes of PWC operation, namely, manual, autonomous, and shared control, when navigating a constrained indoor building environment.
    \item The study demonstrates the technical feasibility and improved usability of the shared control navigation framework in relation to fully manual and autonomous modes based on empirically obtained objective and subjective measures of usability (efficiency, safety, and ease of use).
\end{itemize}

%% file: RelatedWork.tex
\begin{table*}[t]
\centering
\footnotesize
\caption{Comparison between the previous wheelchair models and the CoNav Chair}
\begin{tabular}{ccccc}
\hline
\textbf{Wheelchair Models}&\textbf{SLAM compatible} & \textbf{Autonomous navigation} & \textbf{Shared Control} & \textbf{Sensors}\\
\hline
Powered wheelchair	& N/A & N/A & N/A & N/A \\
VAHM& N/A & N/A & N/A & Contact, Ultrasonic sensor \\
Wheelsley	& N/A & N/A & \cmark & Eye tracking system \\
Vulcan 1.0& \cmark & N/A & N/A & 2D LiDAR and camera \\
Nav-Chair & N/A &\cmark & N/A & 12 ultrasonic sensors \\
Vulcan 2.0	& \cmark & \cmark & N/A & 2D LiDAR \\
Cyberwork & \cmark & \cmark & N/A & 3 Cameras and eye \\
WHILL & N/A & \cmark & N/A & 2D LiDAR \\
\textbf{Co-Nav Chair} (Ours) & \cmark & \cmark & \cmark & 3D LiDAR and RGB-D camera \\
\hline
\end{tabular}
\label{tab: wheelchair comparison}
\end{table*}

\section{Related Work}
\label{sec:relatedwork}


In this section, we review the development of smart wheelchairs and as well as prior work that conducted usability studies on smart wheelchairs.

\subsection{Development of Smart Wheelchairs}

The development of smart wheelchairs has been a focal point of recent literature on improving wheeled mobility technology and independence for PWD. A smart wheelchair typically refers to a powered mobility device equipped with sensors, control systems, and computing hardware that enable advanced functionalities such as obstacle avoidance, autonomous navigation, environmental mapping, and user intent recognition~\cite{Sahoo2023b}. Considerable research on smart wheelchair hardware instrumentation and software integration has been conducted and implemented to address the limitations of traditional PWCs.

Research on smart wheelchairs started in the 1990s with early projects such as VAHM ~\cite{Bourhis1996Mobile}, Vulcan1.0~\cite{Gribble1998Integrating}, Wheelsley~\cite{Yanco1998Wheelesley} and NavChair~\cite{Levine1999NavChair}, continuing on to recent platforms such as the Vulcan2.0~\cite{Park2017DiscreteTime}, Cyberwork Wheelchair~\cite{Kopun2023Toronto} and WHILL~\cite{WHILL}. Table~\ref{tab: wheelchair comparison} compares some of their salient performance capabilities. The VAHM is equipped with a contact sensor, ultrasonic and infrared sensors to detect collisions, and wall-following algorithms to allow the wheelchair to navigate by following the wall. Wheelsley allows wheelchair users to give navigation commands by using an eye-tracking system. Vulcan1.0 uses 2D LiDAR and RGB camera input to build a visual 2D map for the wheelchair user to visualize their current location and surroundings to help improve wheelchair maneuvering. The NavChair is an assistive wheelchair navigation system that integrates autonomous navigation, obstacle avoidance, door passage, and wall-following which allow the wheelchair to navigate autonomously. Vulcan2.0 is the improved version of Vulcan1.0, which integrates autonomous navigation algorithms and reduces costs by using only 2D LiDAR. The Cyberwork Wheelchair is a commercially available smart wheelchair that is equipped with multiple RGB-D cameras to get the surrounding information and use it to navigate autonomously while avoiding obstacles. WHILL introduces an innovative self-driving electric mobility solution, readily available at select airports.

While prior smart wheelchair platforms offer significant navigational assistance to users who have trouble operating traditional PWCs, they have relied entirely on either sensor data or human input, which can have limitations and sensitivity to environmental changes~\cite{Leaman2016DevelopmentOA}. For instance, earlier systems such as VAHM, Wheelsley, Vulcan 1.0, and NavChair are fully manual and require continuous user control, whereas more recent platforms like Vulcan 2.0 and Cyberwork operate in fully autonomous modes without accommodating user intervention. Sensor-driven systems may fail in cluttered or dynamic environments where LiDAR or vision sensors get obstructed or produce noisy data. On the other hand, fully manual systems can overwhelm users when navigating complex indoor environments. User-controlled wheelchair systems require substantial fine motor skills or cognitive effort over extended periods such as those using eye-tracking control~\cite{Yanco1998Wheelesley} which can be gradually fatiguing and unreliable. Fully autonomous systems, on the other hand, can be less reliable in dynamic and unpredictable environments due to sensor limitations or algorithmic shortcomings~\cite{Sahoo2024}. Our proposed system is equipped with multiple sensors and a shared control-based navigation system that allows users to partially control the wheelchair, which increases the reliability of the smart wheelchair system, addressing the limitations of fully autonomous systems in dynamic and unpredictable environments. 

In our prior work, we developed a shared control-based navigation framework that blends user input with autonomous planning to enhance safety and reduce cognitive load in assistive mobility systems~\cite{Xu2024Socially}. Building on this algorithmic foundation, the objective of the present work is to describe the implementation of this smart wheelchair system, and to conduct a preliminary usability evaluation with 21 healthy adults in an operational task environment, namely navigating a narrow corridor in a multistory building, in three different control modes: manual control, fully autonomous, and shared control. Objective (e.g., completion time, trajectory length, collisions) and subjective measures (e.g., self-reported ratings of efficiency, safety, and ease of use) were assessed and compared between the three modes to obtain a comprehensive assessment of the usability of the shared control algorithm in an operational context. 



%% file: Design.tex
\section{Methods}
\subsection{Design of the CoNav Chair}
\label{sec:design}

The CoNav Chair integrates a designed \textbf{hardware} and \textbf{software} system to enable shared control-based navigation for assistive mobility. At the core of the system is a Shared Control-based Model Predictive Control (MPC) framework, which fuses user joystick input with a global navigation plan to generate a dynamic, personalized reference trajectory. This allows users to influence the direction and behavior of the wheelchair based on their intent while relying on the autonomous system for collision avoidance and trajectory optimization. These components shown below establish a tightly coupled perception-control loop that balances autonomy with user control, enabling safe, adaptive, and user-centered navigation in complex environments.

\subsection{Hardware Design}

\par 

\begin{figure}[!tb]
    \centering
    \includegraphics[width =\columnwidth]{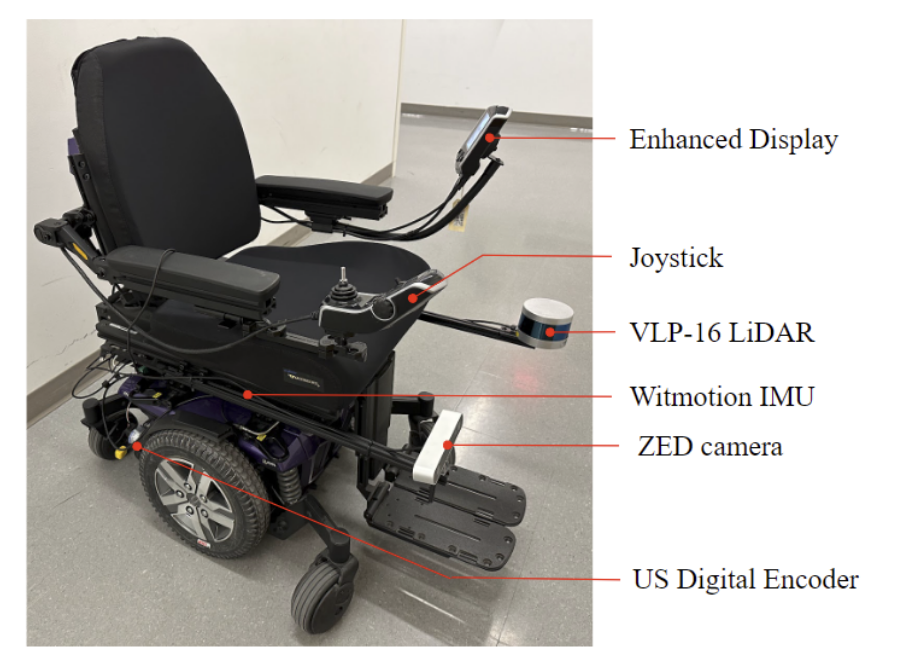}
    \caption{Hardware Design of the CoNav Chair}
    \label{fig:hardware}
\end{figure}

This design is built on a commercially available PWC, Quantum Q6 Edge 2.0, with control electronics, optical encoders, and multiple sensors such as VLP-16 LiDAR, ZED stereo camera, and Witmotion IMU. The hardware assembly is shown in Figure~\ref{fig:hardware}. Sensors like LiDAR and cameras are basic for robot perception, mapping, and autonomous navigation, i.e., simultaneous localization and mapping (SLAM). Unfortunately, most commercial PWC (e.g., Quantum Q6 Edge 2.0, Pride Mobility) are not equipped with the basic sensing system, and some smart wheelchairs in the market are still unaffordable for wide use. Incorporating LiDAR provides precise distance measurement and environmental mapping, enabling accurate obstacle detection and avoidance. The camera adds visual perception, which is essential for recognizing and interpreting complex surroundings and dynamic obstacles. The IMU generates critical data on orientation and movement of the wheelchair to improve stability and control. The encoders contribute to precise wheel movement tracking, enhancing the overall accuracy of the wheelchair's navigation system. Additionally, the closed-loop PID motor control module delivers precise and responsive motor control, which is crucial for smooth and reliable maneuverability. 

\begin{figure}[!tb]
    \centering
    \includegraphics[width =\columnwidth]{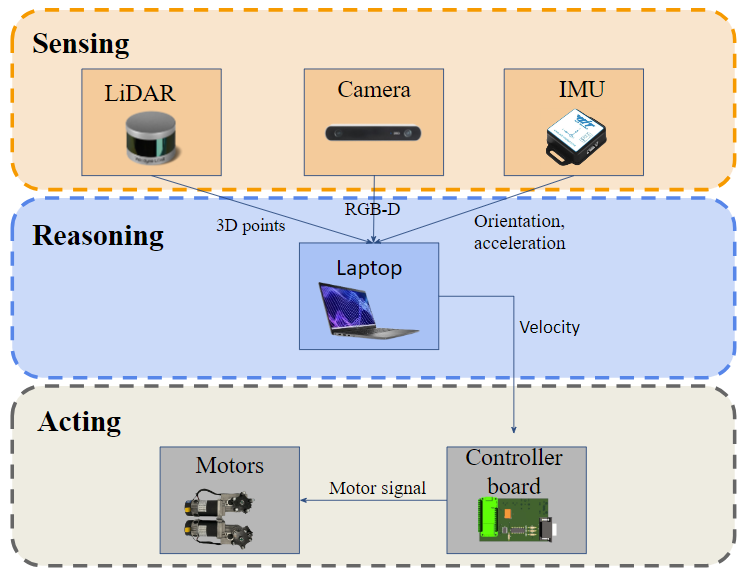}
    \caption{Hardware Dataflow of the CoNav Chair}
    \label{fig:hardwaredata}
\end{figure}

As shown in Figure~\ref{fig:hardwaredata}, our hardware structure follows the Sensing - Reasoning - Acting structure~\cite{Infantino2008}. For the Sensing module, all of the sensors, such as the LiDAR and the camera provide the robot with environmental information about the surroundings and publish sensor data to an onboard laptop computer for further processing. The Reasoning module was supported by a laptop PC connected to the sensors and motors, which run a Linux operating system and ROS. Our shared control-based navigation program runs on this PC and send actuator control commands to motors. For the Acting module, a motor control system based on R-Net is designed, and a closed PID control system is integrated to ensure the stability of the lower-level control system.

\subsubsection{Motion Control Module}

\begin{figure}[!t]
    \centering
    \includegraphics[width =\columnwidth]{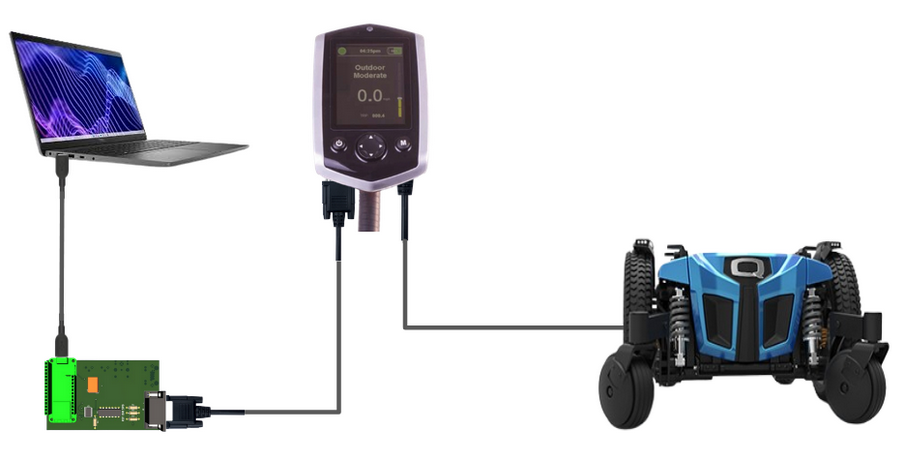}
    \caption{Motor Control Module}
    \label{fig:motor_control}
\end{figure}

For the motor control module, we utilize an alternative control interface that is separate from the standard joystick interface originally used by the powered wheelchair. The alternative control interface is on the enhanced display, which has a 9-way pin-out. We design our motor controller board, which can output two ways of voltage ranging from 4.8V - 7.2V to control both the forward and rotation velocity. The board is programmed to have a \textit{rosserial} module to take command velocity signals sent from our laptop by publishing and subscribing nodes and topic protocol. Inside the board, two digital potentiometers are used to convert the control message to the voltage signal and send it to the alternative control interface. After the alternative control interface gets the voltage output from the controller board, the interface will convert the voltage signals to CAN messages and send them to the wheelchair base. The way we connect the motor controller module is shown in Fig.~\ref{fig:motor_control}.  

\subsubsection{Sensor Module}


The LiDAR and camera were mounted to the wheelchair’s armrest and seat rail using T-slotted aluminum extrusions to avoid obstructing leg movement. The wheelchair motors were modified to retrofit new encoders while retaining the functionality of the integrated electromagnetic brakes. This involved preparing the motor shafts by drilling and threading them to accommodate an encoder mount. The encoders were securely installed using existing mounting points, following standard guidelines, to ensure precise and reliable operation.

\subsection{Software Design}
\par 

\begin{figure*}[h!]
    \centering
    \includegraphics[width =0.8\textwidth]{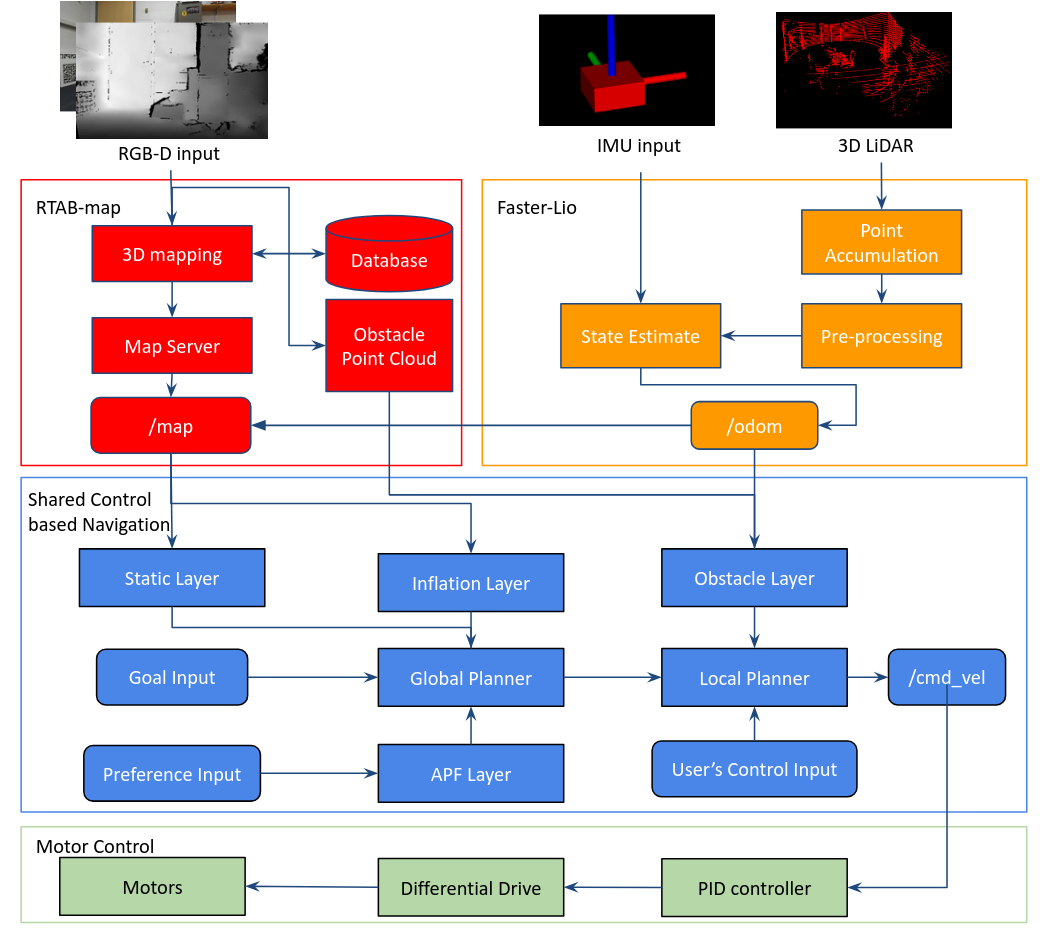}
    \caption{Software Structure of CoNav Chair}
    \label{fig:software}
\end{figure*}

Our proposed SLAM and shared control-based navigation system of the smart wheelchair is based on ROS~\cite{Quigley2009ROS}. ROS provides a robust and flexible framework for writing robot software, allowing developers to leverage a vast ecosystem of tools, libraries, and conventions designed for robot applications. The general structure of the software framework consists of three parts: a 3D SLAM module, a Shared control-based navigation module (Figure~\ref{fig:software}). 

\subsubsection{3D SLAM module}
For the mapping module, the system employs RTAB- Map~\cite{Labbe2013AppearanceBased}, which is a graph-based SLAM algorithm capable of generating both 3D maps and their 2D projections of the surrounding environment. The RGB-D sensor serves as the primary input to RTAB-Map, enabling the construction of detailed 3D maps through real-time point cloud data processing. This map data is stored in a database, which is also used for managing obstacle point clouds for effective collision avoidance. The RTAB-Map framework integrates with a map server to publish the map information, accessible under the \texttt{/map} topic, which is crucial for high-level path planning and navigation tasks.

Simultaneously, the system utilizes Faster-LIO~\cite{Bai2022FasterLIO}, a LiDAR-inertial odometry framework, for accurate localization. Faster-LIO processes data from the 3D LiDAR and IMU inputs. Initially, the raw LiDAR data undergoes pre-processing to filter out noise and structure the point cloud for efficient processing. These pre-processed point clouds are then accumulated to provide consistent spatial information. The state estimation module combines the IMU data with the processed LiDAR information, ensuring robust and precise localization even in dynamic environments. The resulting odometry data is published under the \texttt{/odom} topic, providing real-time position and orientation feedback to the wheelchair system.

\subsubsection{Shared Control-based Navigation}
\begin{figure}[h!]
    \centering
    \includegraphics[width = 0.8\columnwidth]{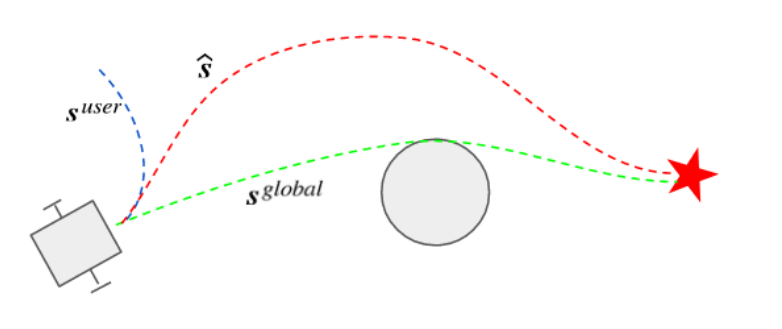}
    \caption{Shared Control-based MPC. Blue line represents user-defined path. Green line represent the path generated by the global planner. Red line represents the path combined.}
    \label{fig:share control}
\end{figure}

By using the predefined map, the wheelchair user can set the desired location for the system to generate the path and navigate the wheelchair to its final destination. During the navigation process, we implement a shared control-based navigation structure, in which the wheelchair user can customize the global path in the user preference-based global planner and adaptively apply their control on the navigation process. This integration allows the system to combine the user's control input signals with the sensor data, enabling the wheelchair to navigate through the environment efficiently and safely. The shared control approach ensures that the user retains control over the wheelchair while benefiting from the system's autonomous navigation capabilities. This balance enhances user trust and promotes a seamless interaction between the user and the wheelchair.

Considering a wheelchair’s role as an intimately user-integrated robotic system, it becomes essential to incorporate considerations of the user’s control. Therefore, within our current framework, we have incorporated methods and concepts of shared control. In order to combine both user’s control and autonomous navigation, we proposed a Shared Control-based Model Predictive Control local planner~\cite{Xu2024Socially} to allow users to adaptively combine their control signals into the whole navigation process. 

Here we utilized a non-linear model predictive control (MPC)~\cite{Wang2021GroupBased} formulation for indoor environments navigation. The optimization problem is formulated below:

\begin{equation}
    \label{eq:vinalla}
    \begin{aligned}
    \mathbf{u}^{*} = \arg&\min_{\mathbf{u}\in\mathcal{U}}\mathcal{J}(\mathbf{s}, \hat{\mathbf{s}}, \mathbf{u})\\
    s.t. s_{t+1} &= f(s_{t}, u_{t}) \\
    \mathbf{s} &\in \mathcal{S}_{free} \\
    \mathbf{u} &\in \mathcal{U}
    \end{aligned}
\end{equation}

where $s = \{s_0,...,s_T\}$  is the state sequence of the robot drawn from the feasible set $S_{free}$by passing a control sequence $u=\{u_0,...,u_{T-1}\}$  drawn from a control space U through the robot dynamic $f$. $\hat{s}=\{\hat{s}_0,...,\hat{s}_T\}$ is the combined reference states of robot by considering both the user's control signal sequence and the global path plan generated by the user-preference field. $\mathcal{J}$ is the cost function that expresses the considerations of efficiency, safety, and user control. 

To enable the user to take control of the navigation process, we incorporate the predicted state generated by the user's control sequence into the cost function $\mathcal{J}$, as shown below.

\begin{equation}
    \label{eq: shared MPC}
    \begin{aligned}
        \mathcal{J}(\mathbf{s}, \hat{\mathbf{s}}, \mathbf{u}) &= \mathcal{J}_s(\mathbf{s}, \hat{\mathbf{s}}) + \mathcal{J}_{u}(\mathbf{u}) \\
        \mathcal{J}_s(\mathbf{s}, \hat{\mathbf{s}}) &= (\mathbf{s}-\hat{\mathbf{s}})^{T}\mathbf{Q}_{s}(\mathbf{s}-\hat{\mathbf{s}}) \\
        \hat{\mathbf{s}} = \eta(k)&\mathbf{s}^{user} + (1-\eta(k))\mathbf{s}^{global} \\
        \mathcal{J}^{u} &= \mathbf{u}^{T}\mathbf{Q}_{u}\mathbf{u}
    \end{aligned}
\end{equation}

where $\mathcal{J}_s$ represents the state cost and $\mathcal{J}_u$ represents the input cost. The combined cost, denoted as $\hat{s}$, takes into account both the global planning state sequence $\mathbf{s}_{global}$ and the user’s control sequence $\mathbf{s}_{user}$. The parameter $\theta(k) \in [0, 1]$ acts as a weight for the user’s control sequence, determining the degree of control the user exerts. The variable k indicates the number of control signals the user provides within a given time frame. To adjust the levels of user control in our system, we devised a weight function based on the frequency of user control, reflecting their intent to steer the wheelchair. An exponential function is used to model the user's intent, as shown below:

\begin{equation}
    \label{eq: shared MPC}
    \begin{aligned}
        \theta(k)=1-e^k
    \end{aligned}
\end{equation}

As shown in Fig.~\ref{fig:share control}, the reference path combined the user’s control signal and the global path plan. The MPC local planner will then allow the robot to follow the combined path while avoiding obstacles.

%% file: Usability.tex
\subsection{Usability Study Design}
\label{sec:usability}

A human subjects usability study with a repeated measures design was conducted to validate the efficiency, safety, and ease of using the proposed shared control framework relative to fully manual and fully autonomous modes of operation. In order to demonstrate the safety of the CoNav wheelchair in an operational setting, this initial study intentionally focused on recruiting healthy, unimpaired adults in the evaluation as a precursor to future evaluations with actual PWC users with impairments. The study was approved by the University of Michigan Institutional Review Board. Study participants provided written informed consent prior to participation.

\subsubsection{Study Participants}
The study recruited a convenience sample of 21 healthy, unimpaired participants for this study. Inclusion criteria for participation included being at least 18 years of age, weighing less than 450 lbs., and not having any previously diagnosed cognitive health condition or upper extremity injury/disorder that could impede the use of a joystick control. The sample size was determined based on previous studies demonstrating that 10-15 individuals were adequate to test for differences in wheelchair use and differentiate usability of a smart wheelchair application \cite{Mantha2020}. Recruited participants in the sample were between the ages of 18 and 45 and with 10 participants identified as male and 11 as female. 

\subsubsection{Study Setting}
\begin{figure}[t]
    \centering
    \includegraphics[width =\columnwidth]{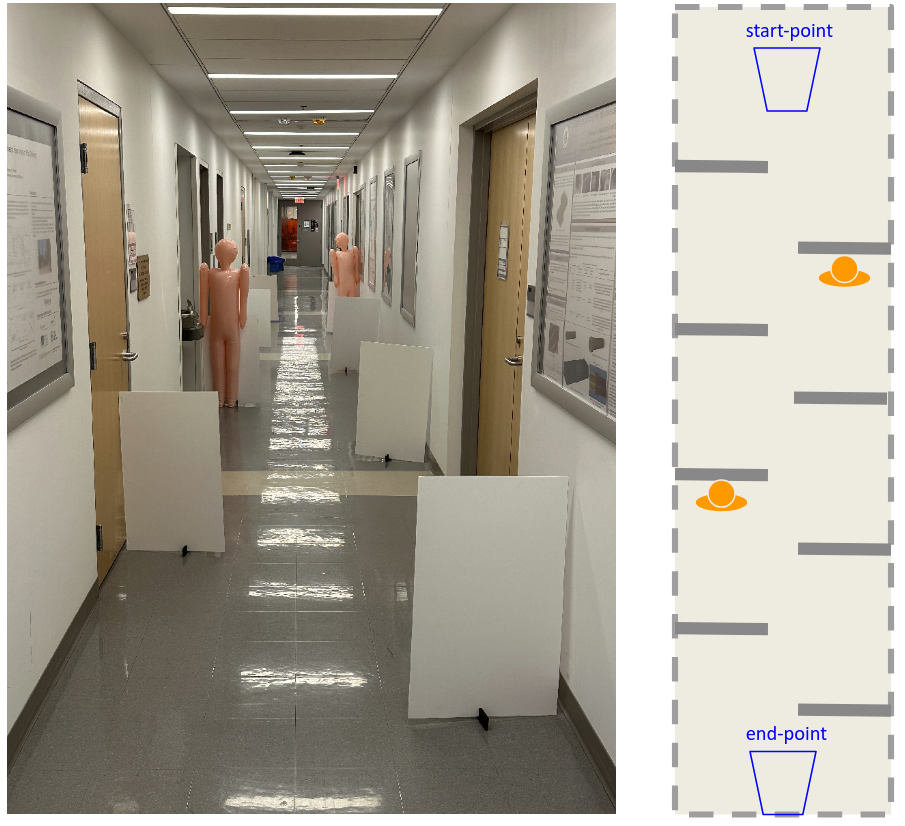}
    \caption{Experiment Environment of Usability Study}
    \label{fig:experiment environment}
\end{figure}

A university building served as the setting for this study. The test environment consisted of a narrow indoor corridor that replicated a challenging real-world context (Figure~\ref{fig:experiment environment}). Eight strategically placed whiteboards served as static obstacles and were arranged alternately along both sides of the corridor to create a zigzag navigation path. In addition, two inflatable human manikins were placed within the environment to simulate pedestrians in the hallway and confirm that the wheelchair could detect different types of barriers. Participants were tasked with traversing the obstacle course twice in each of the three distinct navigation modes moving from a designated start location to a specified end point at a straight-line distance of about 25 m away. The different types of obstacles and spatial constraints in the corridor were intended to evaluate and compare the performance of the three navigation modes in realistic yet controlled conditions.

 \subsubsection{Study Procedure}

The study procedure, including informed consent, required approximately 80 minutes of the participant’s time and consisted of two phases: (1) Introduction and preparation, and (2) Movement trials in the three navigation modes. 


\subsubsection{Phase I: Introduction and Preparation}


First, participants received an introduction to the study objectives, procedure and instructions on wheelchair use.
A brief introduction video was played on a computer explaining the objective of the study, current challenges experienced by wheelchair users, an overview of the proposed shared control-based smart wheelchair, and the study procedure. Next, a brief questionnaire consisting of three questions was administered to the participant to gauge their familiarity with wheelchairs and opinions about the operating of PWCs (about 5 minutes): (1) Have you ever driven a wheelchair before? (2) Have your friends or family ever driven a wheelchair before? (3) What do you think is the biggest challenge when navigating the corridor in a wheelchair?

\subsubsection{Phase II: Movement trials in the three navigation modes}
\begin{figure}[t]
    \centering
    \includegraphics[width =0.8\columnwidth]{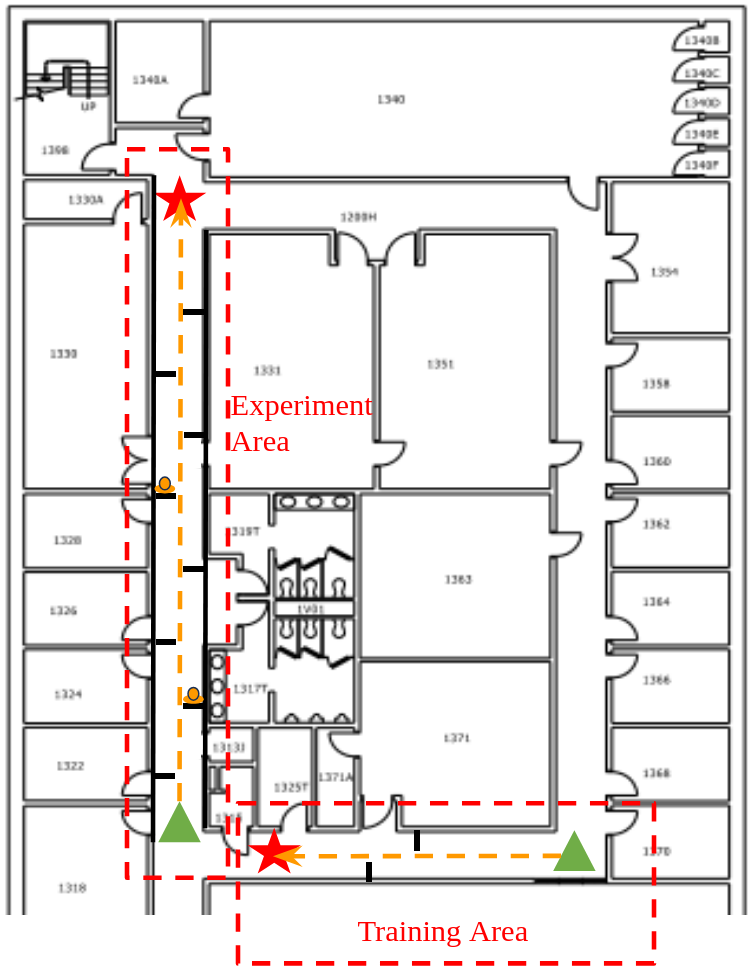}
    \caption{Illustration of the Navigation Path}
    \label{fig:navigation path}
\end{figure}


\begin{figure}[t]
    \centering
    \includegraphics[width =\columnwidth]{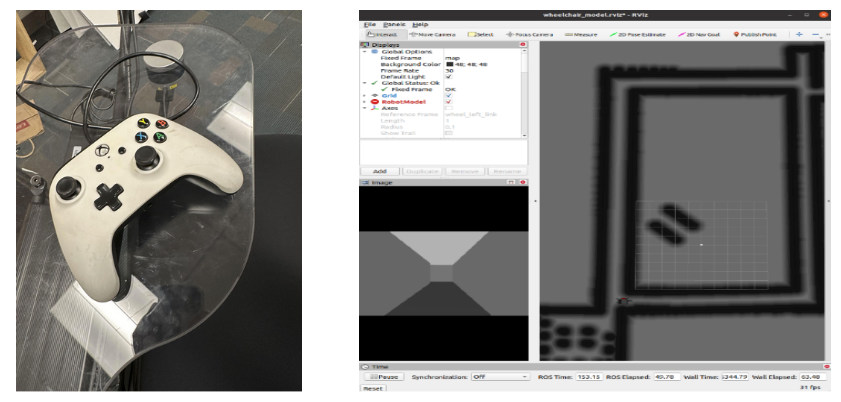}
    \caption{Control Interface for the manual mode (left panel) and autonomous mode (right panel).}
    \label{fig:control interface}
\end{figure}

After completion of Phase I, movement trials were conducted in each of the three navigation modes. The corridor was divided into training and experiment areas as shown in Figure~\ref{fig:navigation path}. To acclimate the participant to each navigation mode and minimize learning effects, the training area was first used to instruct the participant on how to operate the wheelchair and navigate the corridor using the corresponding mode. Upon training and practice, the participant moved to the experiment area to perform the recorded movement trial. For this, the participant was tasked with traversing the dotted path from the start point to the end point by navigating around the obstacles (i.e., whiteboard, inflatable manikins) as quickly and safely as possible avoiding collisions with any obstacle. Two repetitions of the movement trial were performed in each navigation mode.  In each repetition, the participant navigated the wheelchair from the same start point to end point locations for consistency. The order of the navigation modes was kept constant for all participants, with the fully manual mode performed first, autonomous mode second, and shared control mode performed last. The control interface and user interactions in each of the three modes are described below. 

\begin{itemize}
    \item \textbf{Manual Control}: The participant was shown the start and destination location in the floor plan of Fig.~\ref{fig:navigation path}. The participant used an XBox joystick controller (Figure~\ref{fig:control interface} left panel) to manually operate the wheelchair and carefully navigate the obstacle course avoiding any collisions.
    \item \textbf{Autonomous Control}: The participant was shown the start and destination locations on a visual user interface (UI; Figure~\ref{fig:control interface} right panel) and instructed during training on how to set the destination for the robot to navigate autonomously using MPC control\cite{Tahirovic2010}. The participant sat in the wheelchair and hence could experience automated movements and observe system performance firsthand, including how the wheelchair negotiated obstacles by planning and adjusting its path in real-time. 
    \item \textbf{Shared Control}: In the shared control mode, both joystick (manual control) and UI (autonomous control) were used. The participant was reminded of the concept of the proposed shared control-based local planner, wherein after selecting the goal/destination for autonomous navigation, they could use the joystick to influence the speed and direction of the wheelchair at any time of their choosing even while the wheelchair was under autonomous navigation control.
\end{itemize}

A member of the research team followed the participant during each movement at a safe distance and took notes on aspects of system performance, including any observed stops, recalculations, or collisions with obstacles. After two repetitions of the movement trial in each navigation mode, the participant completed a post-trial questionnaire to record their perceptions about that mode. The questionnaire had three sections (with three statements each) corresponding to different aspects of usability, namely ease of use, safety, and efficiency (Table~\ref{tab:post-trial questionnaire}). Participants rated each statement on a 5-point agreement scale, with 1 corresponding to the lowest and 5 corresponding to the highest level of agreement. After evaluating all three navigation modes and associated questionnaires, a brief exit interview was conducted to obtain participant's feedback about the three different modes, potential limitations and provide suggestions for improvement. 

\begin{table*}[h!]
    \centering
    \begin{tabular}{lp{10cm}}
        \hline
        \textbf{Metrics} & \textbf{Questions} \\
        \hline
        \multirow{3}{*}{Efficiency} & 1. I can navigate through your intended path quickly using this mode. \\
        & 2. This mode can help you reduce the time needed to navigate through complex environments. \\
        & 3. This mode helps you maintain a smooth and continuous flow during navigation. \\
        \hline
        \multirow{3}{*}{Safety} & 1. I feel safe navigating in this mode. \\
        & 2. This mode can help you avoid obstacles and potential hazards. \\
        & 3. This mode can handle unexpected obstacles or changes in the environment. \\
        \hline
        \multirow{3}{*}{Ease of Use} & 1. It is easy for me to navigate using this mode. \\
        & 2. I would think people will learn to use this control mode very quickly. \\
        & 3. I don't need to learn too much before I can use this mode. \\
        \hline
    \end{tabular}
    \caption{Post-Trial questionnaire with nine items related to system usability rated for agreement by participants on a 5-point scale.}
    \label{tab:post-trial questionnaire}
\end{table*}

\subsubsection{Dependent Measures}

Separate objective and subjective measures of usability (efficiency, safety, and ease of use) were obtained from the recorded movement trials and subsequently analyzed. The objective measures included the following.
\begin{itemize}
    \item \textbf{Completion Time (seconds)}: The time duration for navigating from the start to the end point was recorded on the onboard computer as a measure of efficiency. Longer time implied lower navigational efficiency. 
    \item \textbf{Trajectory Length (meters)}: The total length of the wheelchair path traveled from the start point to the end point provided an additional measure of efficiency. Longer length implied lower navigational efficiency.
    \item \textbf{Mean Cumulative Angle Difference (radians)}: The cumulative angle difference is a measure of the smoothness of a trajectory \cite{Xu2024Socially} and computed over the entire path traveled by the wheelchair using Equation (4) below and then averaged. A lower average value indicated a smoother travel path.
\begin{equation}
    \mathbf{\theta} = \sum_{i=1}^{n}|\theta_i - \theta_{i-1}|
\end{equation}
where $\mathbf{\theta}$ is the sum of the angle difference and $\theta_i$ and $\theta_{i-1}$ represents the heading of the robot at time step $i$ and $i-1$.

\item \textbf{Control percentage (percent)}: Control percentage is a metric to measure control burden for users~\cite{Suh2016} based on the rate of control inputs to the joystick reflecting both ease of use and safety. A higher value of the control percentage meant a higher burden on the user. Only the manual and shared control modes required use of the joystick and hence were compared on this measure. Participants did not to use the joystick to control the wheelchair in the autonomous control mode and hence was excluded from this measure. 

\item \textbf{Number of Collisions (count)}: Number of collisions during the entire navigation process were recorded as a measure safety. A higher number of collisions meant lower safety. 
\end{itemize}

Subjective measures consisted of user-reported ratings on a 5-point scale to the nine questionnaire statements in three categories (Table~\ref{tab:post-trial questionnaire}), namely: 
\begin{itemize} 
\item \textbf{Efficiency}: reflecting participants' perceptions about the speed and smoothness with which they completed the entire navigation task. 
\item \textbf{Ease of Use}: reflecting the ease with which participants thought they could learn and become proficient in that navigation mode. 
\item \textbf{Safety}: capturing participants' perceived sense of safety while using each of the navigation modes. 
\end{itemize}

\subsubsection{Statistical Analysis}
Statistical data analysis was performed using SPSS v29 (IBM SPSS Statistics Inc.). Data from all 21 participants were used in the analysis, including two repetitions of objective measures and one repetition of subjective assessments. In 9 movement trials (2 manual, 2 autonomous, 5 shared-control modes) participants stopped just before the end location (between 20m and 24m) due to a collision with the obstacle, and hence these 7 trials were excluded from the analysis of completion time and trajectory to reduce measurement bias, but were retained for other measures including cumulative angle difference, number of collisions, and self-reported ratings. Descriptive statistics, including means and standard deviation (SD) for continuous measures, and medians for ordinal and categorical variables, were computed and stratified by navigation mode (manual, autonomous, shared control). Statistical differences between navigation modes were tested using linear mixed-effects models for continuous measures (completion time, trajectory length, cumulative angle difference, and control percentage) to its robustness in handling data with repeated measures, unbalanced samples and moderate sample sizes, providing a more flexible approach compared to traditional methods like ANOVA. The number of collisions and self-reported ratings to the usability questions were not normally distributed, and hence tests for differences by navigation mode were examined using the nonparametric Kruskal-Wallis test. A significance level of $p < .05$ was used for all tests. Significant main effects of navigation mode were subsequently examined using pairwise adjusting for multiple comparisons using the Bonferroni procedure ($p < .05$). 

%% file: Result.tex
\section{RESULTS}
\label{sec:result}

\subsection{Pre-Trial Survey}


Responses from the pre-test survey on participants’ prior experience of wheelchairs indicated that 95.2\% of participants had no experience in using wheelchairs, and 81.0\% did not have friends and family members or friends that ever used a wheelchair before. These results suggest that most participants approached the experiment with limited firsthand or secondhand exposure to wheelchair mobility. As a preliminary study, this lack of prior experience implied that the participants had a similar baseline knowledge about wheelchairs and were less likely to have preconceived preferences, which served to reduce bias when comparing the three navigation modes and allowing for a more objective evaluation of usability, safety, and efficiency. 

The most challenging aspects of wheelchair navigation reported by participants were obstacles (e.g., furniture, people or objects) (N = 15; 71\%), followed by difficulty in maneuvering turns (N = 13; 62\%), narrow pathways (N = 7; 33\%), and uneven or slippery flooring (N = 4; 19\%). This finding underscored common concerns among the study participants about maneuvering through cluttered or narrow indoor spaces and provided some support to the validity of our experimental setup. Reflecting the perceived real-world challenges of participants in the test environment allowed for more meaningful and realistic evaluation of the different control modes.

\subsubsection{Objective Measures of Usability}

\begin{figure}[t]
    \centering
    \includegraphics[width =\columnwidth]{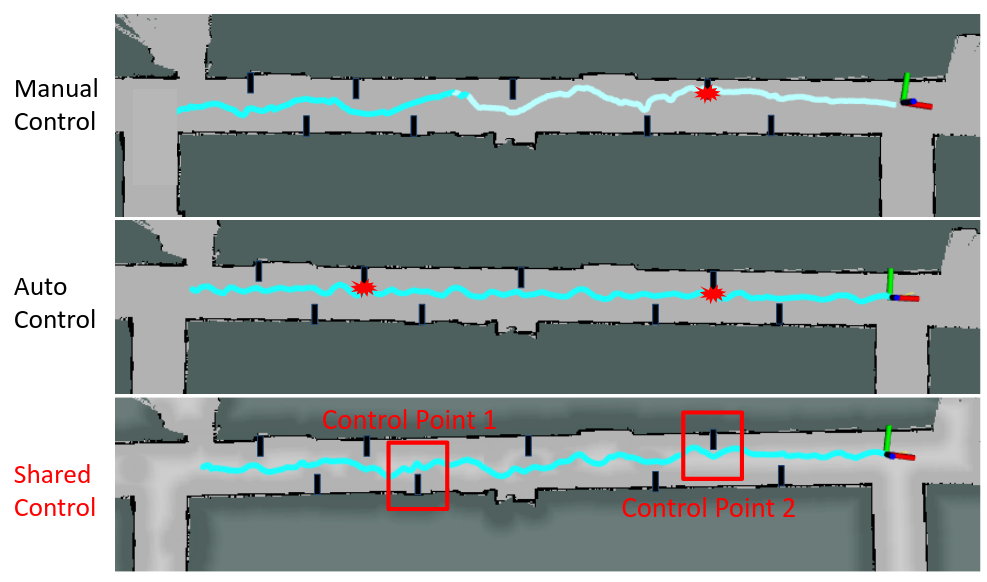}
    \caption{Trajectory Comparison Across Navigation Modes}
    \label{fig:traj compare}
\end{figure}

From the trajectory comparison shown in Fig.~\ref{fig:traj compare}, it is clear that shared control effectively combines human intuition and algorithmic decision-making to avoid collisions, while manual and autonomous controls each face distinct challenges. In manual control, the subject took a wide turn when encountering obstacles but ultimately collided due to inexperience and the high cognitive burden of constant control. In contrast, the autonomous mode—driven by model predictive control (MPC)—attempted to optimize for the shortest route, leading to potentially risky paths and resulting in two collisions. By integrating the user’s judgment with the algorithm’s path-planning capabilities, shared control avoids both the inexperience-driven errors of manual navigation and the over-optimized paths of fully autonomous navigation, achieving a collision-free trajectory. The following is a more detailed analysis of the measures extracted from these wheelchair trajectories.

\begin{table*}[htbp]
\centering
\resizebox{\textwidth}{!}{
\begin{tabular}{@{}lccc@{}}
\toprule
\textbf{Measure} & \textbf{Manual} & \textbf{Autonomous} & \textbf{Shared-control} \\
\midrule
Completion Time (s) & 92.1 ± 22.6 (67.2 - 183.0) & 72.8 ± 9.2 (60.9 - 115.1) & 72.5 ± 6.3 (51.6 - 81.7) \\
Trajectory Length (m) & 26.0 ± 1.8 (23.8 - 33.1) & 25.6 ± 1.7 (23.3 - 33.4) & 25.3 ± 0.8 (23.6 - 27.0) \\
Cumulative Angle Difference (rad) & 11.2 ± 4.9 (3.4 - 32.1) & 13.5 ± 5.7 (10.0 - 48.1) & 12.9 ± 2.5 (6.2 - 16.5) \\
Control Percentage (\%) & 61.2 ± 14.2 (19.8 - 86.8) & n/a & 14.0 ± 4.8 (4.7 - 23.2) \\
Number of Collisions & 1 ± 1 (0 - 2) & 4 ± 2 (0 - 5) & 0 ± 1 (0 - 2) \\
\bottomrule
\end{tabular}
}
\caption{Descriptive statistics for the objective measures stratified by navigation mode. Values include mean and SD for continuous variables, median and inter-quartile range for ordinal variables (number of collisions), along with the range (minimum - maximum) for all measures.}
\label{tab:quantitative result}
\end{table*}

Table~\ref{tab:quantitative result} provides summary statistics on the objective measures of completion time, trajectory length, cumulative angle difference, control percentage, and number of collisions by navigation mode. Statistics for completion time and trajectory length exclude 9 trials in which participants stopped before the end location.


Mixed effects analysis of completion time revealed significant differences by control mode ($F = 16.96, p < .001$). Pairwise comparisons indicated that Completion times in the shared control and autonomous modes were significantly less than the manual mode ($p < .001$ for both); however the shared control and autonomous modes were not significantly different from each other ($p = 1.0$). 
The analysis of trajectory length indicated a slightly shorter length in the shared control mode compared to the manual and autonomous modes, but these differences were not statistically significant ($F = 2.15, p = .125$). Together, these results indicate that in terms of the efficiency of movement,  the shared control mode was similar, if not marginally better than the autonomous mode, since the autonomous path planning system underlying both these modes is likely already well optimized with limited room for further gains through human intervention. However, the significantly larger reduction compared to manual control suggests that shared control substantially alleviates the cognitive and operational burden on human operators who might otherwise struggle with the complexities of information processing and continuous online control of the wheelchair. By blending human intuition with the automated functions of the robot, shared control reduced some of the inefficiencies in manual operation, such as hesitation or suboptimal path choice, while only marginally improving upon an already efficient autonomous navigation system.

The average cumulative angle difference which served as a measure of the smoothness of the measured wheelchair trajectory, had a lower average value or smoother movement for the manual mode followed by shared control and lastly the autonomous mode; however these differences were relatively small and not statistically significant ($F = 2.06, p = .136$). In this metric, manual control exhibits the smoothest motion because a human operator often makes deliberate, gradual course corrections rather than rapidly oscillating or micro-adjusting, which minimizes abrupt changes in direction and speed. In contrast, the autonomous mode may continuously and reactively adjust its path to optimize for safety or efficiency, leading to frequent small course corrections and thus higher curvature variance. These continuous fine-grained adjustments in autonomous operation reduced overall travel time but caused less smooth trajectories compared to more fluid maneuvering by a human operator. 

The ease of use in interaction with the wheelchair was assessed by comparing the percent control of the wheelchair in the manual and shared control modes (Table~\ref{tab:quantitative result}). A significant difference between both modes was found ($F = 277.7, p < .001$) with the control percentage in the shared control mode being nearly four times less than that of manual control ($p < .001$), implying that the control burden in the shared control mode was far less than than shared control. Since shared control delegates much of the navigation work to autonomous navigation and only requires user intervention for strategic or corrective inputs, the percentage of joystick adjustments is inherently far lower than in fully manual control, where the user must continuously guide the wheelchair. This substantial reduction in required inputs reflects a corresponding decrease in both physical and cognitive load on the user, thereby making shared control significantly less burdensome than manual operation.

System safety was assessed by comparing the number of collisions between all three three modes (Table.~\ref{tab:quantitative result}). The total number of collisions was least in the shared control mode (N = 18 collisions), followed by the manual mode (N = 27) and autonomous mode (N = 158). Nonparametric analyses of median collision counts indicated significant differences between the control modes ($\chi^2  = 89.81, p < .001$). Pairwise comparisons indicated significantly fewer collisions in the shared control vs. autonomous mode (p < .001) and manual vs. autonomous mode ($p < .001$) but not between the shared control vs. manual modes ($p = .126$). The fewer collisions in the shared control compared to autonomous mode can be attributed to the synergy of human oversight and automated assistance. In fully autonomous mode, misinterpretation of rapidly changing environments or system uncertainties might lead to collisions before the system can react. Meanwhile, purely manual control relies solely on the user’s perception and reaction times, which can be prone to human error and fatigue. By combining the strengths of both, shared control allows the human to intervene in ambiguous or critical situations while still benefiting from the consistency and precision in decision-making by the robot. This complementary relationship effectively leverages the strengths of the fully manual and autonomous modes alone, leading to significantly fewer collisions.

\subsubsection{Self-reported Usability Ratings}


\begin{table}[t]
\centering
\resizebox{0.8\columnwidth}{!}{%
\begin{tabular}{@{}lccc@{}}
\toprule
\textbf{Question} & \textbf{Manual} & \textbf{Autonomous} & \textbf{Shared-control} \\
\midrule
Q1 & 3 ± 2 & 4 ± 2 & 4 ± 2 \\
Q2 & 3 ± 2 & 3 ± 2 & 4 ± 2 \\
Q3 & 3 ± 2 & 3 ± 3 & 4 ± 3 \\
Q4 & 4 ± 1 & 2 ± 2 & 4 ± 2 \\
Q5 & 4 ± 1 & 1 ± 2 & 4 ± 1 \\
Q6 & 4 ± 3 & 1 ± 2 & 4 ± 2 \\
Q7 & 3 ± 3 & 4 ± 3 & 4 ± 2 \\
Q8 & 4 ± 2 & 5 ± 1 & 4 ± 2 \\
Q9 & 5 ± 2 & 5 ± 0 & 4 ± 1 \\
\bottomrule
\end{tabular}%
}
\caption{Median and interquartile range values by navigation mode for participant ratings to the nine-item questionnaire described in Table~\ref{tab:post-trial questionnaire}.}
\label{tab:questionnaire_ratings}
\end{table}


Overall, the shared control model scored a median rating of 4/5 on all nine statements presented to participants while median ratings for manual and autonomous modes varied between 1 and 5 (Table.~\ref{tab:questionnaire_ratings}). Nonparametric analysis of the nine questions after the two repetitions of movement trials indicated important trends. For the first three questions related to ease of use, median ratings for the shared control mode were slightly higher than the other two modes. Specifically, ratings differed significantly in response to Q2 (\textit{I would think people will learn to use this control mode very quickly}; $\chi^2  = 6.22, p < .045$). Pairwise comparisons indicated that the shared control mode was rated significantly easier to use than the autonomous mode ($p = .030$). However, no significant differences in mode were found for ratings associated with Q1 (\textit{I don't need to learn much before I can use this mode}; $\chi^2  = 2.18, p < .337$) and Q3 (\textit{It is easy for me to navigate using this mode}; $\chi^2  = 4.34, p = .114$). These findings likely reflects the reduced user workload in autonomous mode, where the system handles most of the decision-making and movement, requiring users only to select a navigation goal. In contrast, shared control relies on occasional user intervention for course corrections, while manual control requires continuous input, making it more demanding for users.

For the next three questions related to perceived safety (Q4 - Q6), median ratings were higher, implying safer in the shared control and manual modes compared to the autonomous mode. Significant differences in median ratings by mode were obtained for all three questions. Specifically, for Q4 (\textit{This mode can handle unexpected obstacles or changes in the environment}; $\chi^2  = 10.69, p = .005$), the shared control and manual modes were similar ($p = .05$), though rated higher compared to autonomous mode ($p = .002$ and $p = .011$, respectively). Median ratings for Q5 (\textit{This mode can help you avoid obstacles and potential hazards}; $\chi^2  = 6.01, p = .050$), followed a similar trend with the shared control and manual modes rated similar ($p = .079$), though rated higher compared to autonomous mode (p < .001 for both comparisons). Likewise, for Q6 (\textit{I feel safe navigating in this mode}; $\chi^2  = 17.76, p < .001$), the shared control and manual modes were rated similar ($p = .513$) and both rated higher compared to the autonomous mode ($p < .001$ for both comparisons). These findings suggested that users appreciate retaining some control over wheelchair movement, as it provides both the ability to intervene when needed and the reassurance of autonomous assistance. By offloading tasks such as obstacle detection and path planning to the system, shared control reduces the cognitive and perceptual burden on users, which may help prevent accidents and increase overall confidence in safety.

The last three survey questions (Q7 - Q9) focused on perceived efficiency. Analysis indicated significant effects of mode for Q9 (\textit{I can navigate through your intended path quickly using this mode}; $\chi^2  = 9.23, p < .0105$), but not Q7 (\textit{This mode helps you maintain a smooth and continuous flow during navigation}; $\chi^2  = 2.65, p < .265$) and Q8 (\textit{This mode can help you reduce the time needed to navigate through complex environments}; $\chi^2  = 4.91, p < .0865$). Pairwise comparisons for Q9 indicated that the autonomous mode was rated higher than the shared control mode ($p = .008$) and manual mode ($p = .008$). Generally, participants felt that shared control and autonomous modes were more efficient and marginally outperformed the manual mode. This likely stems from the combination of machine-driven optimizations, which help minimize both unnecessary detours and collisions. Furthermore, users perceived that they could accomplish navigation tasks more quickly and reliably when they collaborate with the MPC autonomous navigation algorithm rather than relying solely on fully manual control.


\subsubsection{Reflections from Post-trial Interviews}


Important insights were also obtained from participants' responses to the three open-ended interview questions asked after they completed movement trials in all three modes. Key themes are summarized below:

\begin{itemize}
    \item \textit{Which navigation mode is your favorite? Manual Control, Autonomous Control, or Share Control?}

In response to this question, 15 participants stated a preference for shared control, six participants preferred manual control the most,  and none of the participants preferred the fully autonomous mode. We have noted that participants who had used a joystick or a wheelchair before were more likely to have a preference for manual control instead of the other two modes. This suggests that prior experience might influence user preferences, while those lacking extensive joystick experience may find shared control more intuitive and reassuring than purely manual modes. 
    
    \item \textit{How confidently can you use the shared control after the experiment? }

Participants were asked to rate how confident they were in using the shared control on the wheelchair with a 5-point likely scale rating. The shared control mode received a median confidence rating of 5 out of 5. Specifically, the participants who rated 5 out of 5 stated that they felt it was both intuitive and suitable for daily use. The other 10 participants gave it 4/5, explaining that the shared control mode occasionally reduced navigation smoothness, prompting them to deduct one rating point.

    \item \textit{What are the limitations of our proposed shared control-based wheelchair?}

Several participants noted that the system exhibited a noticeable delay between their joystick inputs and the wheelchair’s response. As one participant described, \textit{"It felt like the system was a bit slow to react, so I ended up pushing the joystick more just to correct the direction."} Others reported that this led to over-corrections: \textit{"By the time it responded, I had already adjusted again, and then it would suddenly jerk too much."} This feedback suggests that users may have experienced some response latency that influenced the intended smoothness and synergy of the shared control experience, indicating a need for further system improvements.

\end{itemize}

\subsection{Discussion}
This study presented and evaluated a novel shared control-based smart wheelchair platform called the CoNav Chair that bridges the gap between full autonomy and manual control, and thereby advances assistive mobility solutions. Findings from a controlled study with 21 healthy participants without impairment helped to demonstrate the technical feasibility, performance and improved usability of the shared control navigation framework in relation to fully manual and autonomous modes based on empirically obtained objective and subjective measures of usability (efficiency, safety, and ease of use). Specifically, the usability study provided critical insights into how different navigation modes: manual, autonomous, and shared control, perform in a realistic operational environment. Our findings underscore several important lessons for designing assistive navigation systems.

One of the most significant takeaways is that the shared control approach strikes an effective balance between the extremes of fully manual and fully autonomous navigation. While the autonomous mode requires minimal user intervention, it does not allow any room for corrective action during the navigation process. In contrast, manual control demands constant user input, increasing both cognitive and physical load. The shared control mode, however, integrates user intervention with autonomous decision-making, allowing operators to step in when necessary. This not only reduced the number of collisions significantly but also served to build user trust by maintaining a sense of control.

Findings from the objective measurements indicated that shared control and autonomous navigation modes shared similar task time and trajectory length. Manual control, on the other hand, consistently underperformed in efficiency due to its higher sensorimotor and cognitive demands on the user. However, the shared control system effectively minimizes unnecessary detours and reduces the operational workload by offloading complex tasks—such as obstacle detection and path optimization—to the autonomous subsystem. This synergy results in navigation performance that is both reliable and efficient, making it a viable solution for real-world indoor environments.

The subjective assessments reveal that participants feel most secure when using shared control. Although autonomous navigation was rated as the easiest to use, the added ability to intervene when necessary gave participants a heightened sense of safety. Furthermore, the data suggest that prior experience with joystick-based control influences user preference. Participants with previous experience tended to lean towards manual control, whereas those without such experience found shared control more intuitive and reassuring. This highlights the importance of user-centered design and the need to tailor control schemes to diverse user backgrounds.

In practical terms, these results suggest that the CoNav Chair can substantially improve indoor mobility for users with diverse skill levels and needs. By fusing user-driven inputs with an autonomous navigation algorithm, the system not only mitigates safety concerns but also boosts navigation efficiency and comfort. From a deployment perspective, healthcare facilities, rehabilitation centers, and assisted living environments could integrate this shared-control approach to help individuals with varying degrees of mobility challenges navigate crowded corridors or dynamic indoor spaces.

\subsection{Study Limitations}
Despite its benefits, the shared control system exhibited some limitations. A notable issue was system delay—the lag between user input and system response occasionally led to overcorrections, which disrupted the intended synergy between human and autonomous control. Importantly, this delay primarily stems from limitations in the wheelchair's hardware design rather than the shared control framework itself. We have implemented various compensation strategies, including PID and Model Predictive Control (MPC), to mitigate the effect of this latency. Addressing this latency would be essential to improve both navigation fluidity and overall user satisfaction. 
In order to improve the shared control system, improving the hardware system and reducing the system delay is key. In addition, we would like to explore more about applying the algorithm in dynamic socially-aware environments by deploying our previously proposed SS-MPC-DCBF algorithm~\cite{Xu2024Socially}. 

A limitation of the usability study is that it only had healthy, able-bodied participants. However, as an initial study with unproven performance and safety, recruiting such healthy participants was considered both essential and ethical, before conducting subsequent human subject evaluations with actual end users, namely individuals with mobility disabilities who use PWCs. Lastly, exploring additional interaction modalities (such as voice control or gesture recognition) alongside traditional joystick inputs could offer users more flexible control options, particularly for users with limited upper extremity strength and dexterity who are otherwise unable to safely operate a conventional joystick controller.

%% file: Conclusion.tex
\section{Conclusion}
\label{sec:conclusion}
This paper introduced and comprehensively evaluated the CoNav Chair, a shared-control smart wheelchair designed to address the limitations of fully manual or fully autonomous navigation in an indoor operational environment with obstacles. The study involved 21 participants navigating a narrow corridor in three different modes—manual, autonomous, and shared control—so that we could assess the system’s usability using both objective and subjective measures. The quantitative findings showed that shared control reduced collisions significantly while also alleviating user workload. In subjective assessments, most of the participants rated shared control as offering the best combination of safety, efficiency, and ease of use, especially for those without prior joystick or wheelchair experience.

In terms of practical relevance, these results suggest that the CoNav Chair can substantially improve indoor mobility for users with diverse skill levels and needs. By fusing user-driven inputs with an autonomous navigation algorithm, the system not only mitigates safety concerns but also boosts navigation efficiency and comfort. From a deployment perspective, healthcare facilities, rehabilitation centers, and assisted living environments could integrate this shared-control approach to help individuals with varying degrees of mobility challenges navigate crowded corridors or dynamic indoor spaces.

Further work is needed to optimize the CoNav Chair’s responsiveness to user inputs by mitigating existing system delays and refining the Model Predictive Control (MPC) module for smoother, more adaptable navigation. Future expansions may include integrating advanced sensors for robust obstacle detection, exploring adaptive machine learning techniques to personalize control parameters based on an individual’s experience and preferences, and extending the system’s capabilities to more complex, real-world settings with dynamic human crowds. By pursuing these enhancements, the CoNav Chair could pave the way for a new generation of assistive robotic platforms that balance autonomy and user agency in a wide array of built environments.

%% file: data.tex
\section{Data Availability Statement}
\label{sec:data}
Certain aspects of the code, algorithms, and data supporting the findings of this study are available from the corresponding author upon reasonable request.

%% file: Acknowledge.tex
\section{Acknowledgements}
\label{sec:acknowledge}
The work presented in this paper was supported with funding from the United States National Science Foundation (NSF) through award\# SCC-IRG 2124857. The support of the NSF is gratefully acknowledged.